\documentclass[letterpaper, 10 pt, journal, twoside]{IEEEtran}

\ifCLASSINFOpdf
  \usepackage[pdftex]{graphicx}
\else
\fi

\usepackage{amsmath}
\usepackage{bm}
\usepackage{amssymb}
\usepackage{url}
\usepackage{hyperref}
\hypersetup{
    colorlinks=true,
    linkcolor=magenta,
    filecolor=magenta,      
    urlcolor=magenta,
}
\def\secref#1{Sec.~\ref{#1}}
\def\figref#1{Fig.~\ref{#1}}
\def\tabref#1{Tab.~\ref{#1}}
\def\eqref#1{Eq.~(\ref{#1})}

\usepackage{multirow}
\usepackage{booktabs}

\begin{document}

\title{LCPR: A Multi-Scale Attention-Based LiDAR-Camera Fusion Network for Place Recognition}

\author{Zijie Zhou$^{1}$, Jingyi Xu$^{2}$, Guangming Xiong$^{1}$, and Junyi Ma*
\thanks{Manuscript received: August, 20, 2023; Revised November, 14, 2023; Accepted December, 13, 2023.}
\thanks{This paper was recommended for publication by Javier Civera upon evaluation of the Associate Editor and Reviewers' comments.
This work was supported by the National Natural Science Foundation of China under Grant 52372404.} 
\thanks{$^{1}$Zijie Zhou and Guangming Xiong are with Beijing Institute of Technology, Beijing, 100081, China
        {\tt\footnotesize 3120220445@bit.edu.cn}
        }
\thanks{$^{2}$Jingyi Xu is with Shanghai Jiao Tong University, 200240, Shanghai
        {\tt\footnotesize 3220200359@bit.edu.cn}
        }
\thanks{$^*$Junyi Ma (corresponding author) is with Beijing Institute of Technology and HAOMO.AI Technology Co., Ltd
        {\tt\footnotesize junyi.ma@bit.edu.cn}
        }

\thanks{Digital Object Identifier (DOI): see top of this page.}
}

\markboth{IEEE Robotics and Automation Letters. Preprint Version. Accepted December, 2023}
{ZHOU \MakeLowercase{\textit{et al.}}: LCPR: A Multi-Scale Attention-Based LiDAR-Camera Fusion Network for Place Recognition} 

\maketitle

\begin{abstract}
Place recognition is one of the most crucial modules for autonomous vehicles to identify places that were previously visited in GPS-invalid environments. Sensor fusion is considered an effective method to overcome the weaknesses of individual sensors. In recent years, multimodal place recognition fusing information from multiple sensors has gathered increasing attention. However, most existing multimodal place recognition methods only use limited field-of-view camera images, which leads to an imbalance between features from different modalities and limits the effectiveness of sensor fusion. In this paper, we present a novel neural network named LCPR for robust multimodal place recognition, which fuses LiDAR point clouds with multi-view RGB images to generate discriminative and yaw-rotation invariant representations of the environment. A multi-scale attention-based fusion module is proposed to fully exploit the panoramic views from different modalities of the environment and their correlations. We evaluate our method on the nuScenes dataset, and the experimental results show that our method can effectively utilize multi-view camera and LiDAR data to improve the place recognition performance while maintaining strong robustness to viewpoint changes. Our open-source code and pre-trained models are available at \url{https://github.com/ZhouZijie77/LCPR}.\\
\end{abstract}

\begin{IEEEkeywords}
Place Recognition, SLAM, Sensor Fusion, Deep Learning.
\end{IEEEkeywords}

\IEEEpeerreviewmaketitle

\section{Introduction}
\label{sec:intro}

\IEEEPARstart{G}{iven} an observation from sensors such as LiDARs and cameras, the place recognition task needs to find where the observation was previously collected, which is usually used as the first step of loop closure detection in SLAM systems or global localization~\cite{yin2022general, ma2022overlaptransformer}.
Cameras are the most commonly used sensors in various autonomous driving systems, due to their low cost and ability to provide rich semantic information. Vision-based place recognition has been extensively studied, including early approaches based on hand-crafted descriptor~\cite{bay2006surf, tang2019distinctive,torii2013visual} and more recent deep learning-based methods~\cite{arandjelovic2016netvlad, hausler2021patch, khaliq2020camal}. However, features extracted from images are prone to be affected by weather, season, and lighting conditions, especially in outdoor environments, significantly diminishing the localization performance.
Compared to cameras, LiDARs can capture 3D structural information of the environment and are not sensitive to the factors mentioned above, which has been demonstrated by many previous LiDAR-based place recognition works~\cite{kim2018scan, uy2018pointnetvlad, liu2019lpd, ma2022overlaptransformer}. However, the performance of these LiDAR-based approaches still suffers from viewpoint changes, sparsity of point clouds, and lack of texture information.

\begin{figure}
  \centering
  \includegraphics[width=1\linewidth]{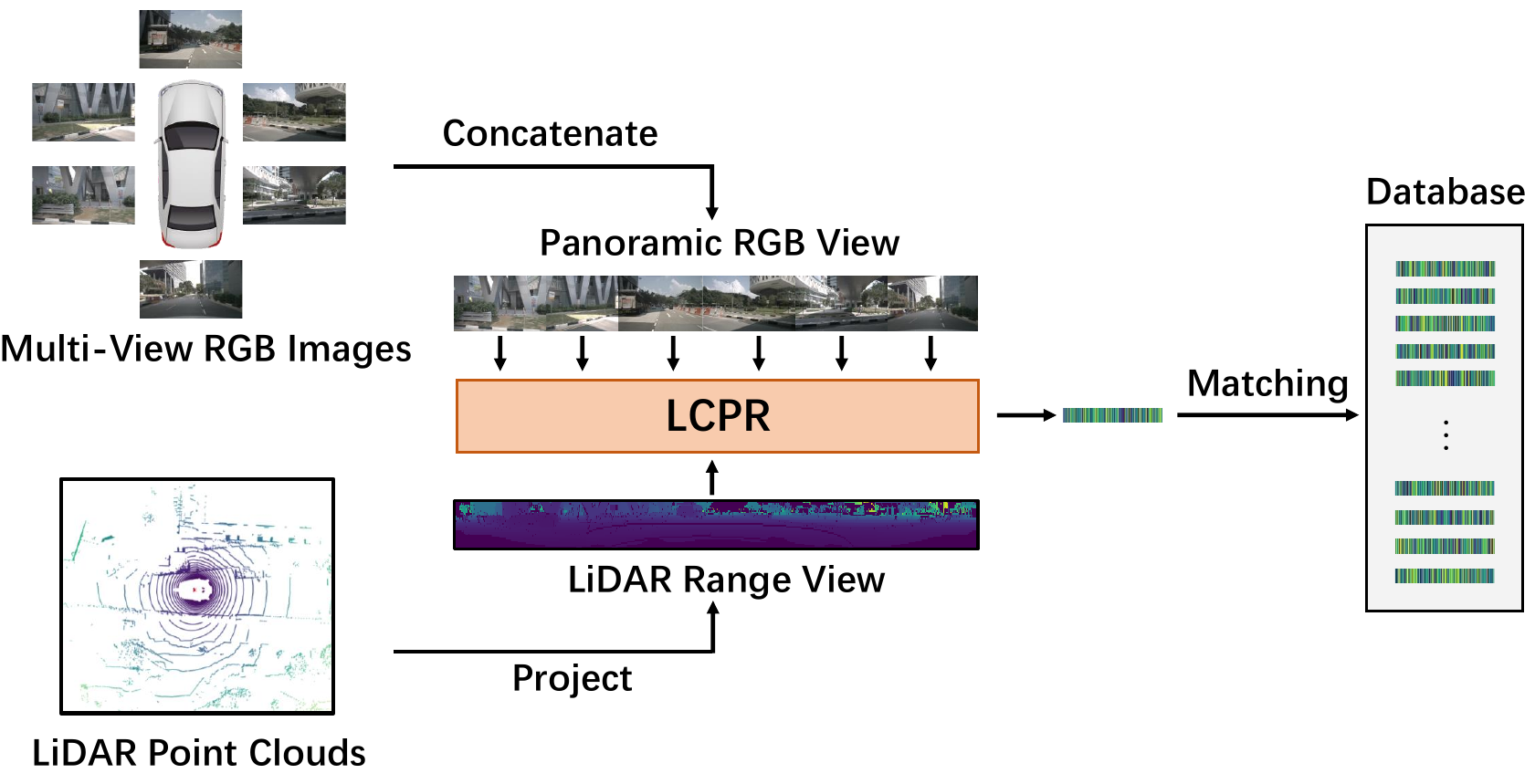}
  \caption{Multi-view images can provide full perspective information of the environment as LiDAR does. LCPR leverages multi-view RGB images and range image from LiDAR as inputs, and utilizes transformer attention to identify the correspondence between two modalities. Localization can be achieved by searching for the nearest neighbor in the database.}
  \label{fig:motivation}
\end{figure}

The combination of LiDARs and cameras can surpass the individual limitations within each modality, which is widely used in autonomous driving perception tasks, such as object detection, semantic segmentation, and depth prediction.
In the field of place recognition, some multimodal fusion-based approaches have been proposed, such as AdaFusion~\cite{lai2022adafusion}, MinkLoc++~\cite{komorowski2021minkloc++}, PIC-Net~\cite{lu2020pic}, proving that multimodal fusion can enhance the performance of recognition. However, the limited field of view (FOV) and viewpoint of the image data used by these methods introduce an inherent flaw when the observation angles are altered by mounting positions or driving directions, resulting in an imbalance between features from different modalities for sensor fusion.

In this paper, inspired by the successful application of multimodal fusion in 3D object detection and map segmentation tasks~\cite{liu2023bevfusion, prakash2021multi}, we propose a multi-scale attention-based network LCPR, as shown in~\figref{fig:motivation}, to fuse multimodal sensor information for place recognition. We focus on extracting discriminative and yaw-rotation invariant global descriptors from both the 3D LiDAR point clouds and RGB images from multi-view cameras for fast query-database matching. 
The environmental features from the two modalities are fused at multiple scales by the proposed Vertically Compressed Transformer Fusion (VCTF) module to fully exploit the panoramic features from LiDARs and cameras as well as their intra- and inter-correlations.

In summary, the main contributions of our work are as follows:
\begin{itemize}
\item We propose LCPR, a multi-scale attention-based network designed to generate discriminative and yaw-rotation invariant multimodal descriptors for place recognition, using 3D LiDAR point clouds and 2D RGB images from multi-view cameras as joint inputs.
\item A novel multimodal fusion framework and a panoramic fusion module are proposed, utilizing self-attention mechanism of transformers to identify and exploit the correspondences between panoramic features from different modalities.
\item We evaluate our proposed LCPR on the nuScenes dataset~\cite{caesar2020nuscenes}, and the experimental results validate that our method outperforms the current state-of-the-art methods including both the unimodal and multimodal ones.

\end{itemize}

\section{Related Work}
\label{sec:related}

The performance of place recognition is strongly related to the adaptability of various types of sensors. Therefore, we make a brief review of place recognition methods based on different sensor setups.

\subsection{Vision-based Place Recognition}
\label{sec:VPR}
Vision-based place recognition (VPR) is usually regarded as an image retrieval problem, where the database image most similar to the current query image is recalled. Traditional image retrieval methods are proposed to extract hand-crafted local feature descriptors.
For example, Andreasson~\cite{andreasson2004topological} uses the simplified SIFT~\cite{lowe2004distinctive} for topological localization for mobile robots. After that, Zhang \textit{et al.}~\cite{zhang2006image} introduce a novel robust motion estimation technique to deal with a larger percentage of mismatches based on SIFT features.
In~\cite{jegou2010aggregating}, Jégou \textit{et al.} introduce the VLAD descriptor for compact image representation, which shows effectiveness and efficiency in large scale image retrieval tasks.
With the rapid development of deep learning, researchers are now more interested in using neural networks to extract high-level place features instead of the hand-crafted ways.
In~\cite{chen2014convolutional}, Chen \textit{et al.} first introduce CNNs into the place recognition system. Arandjelović \textit{et al.}~\cite{arandjelovic2016netvlad} propose NetVLAD, a feature aggregation layer which can be easily plugged into any CNN architecture and trained in an end-to-end manner. After that, Radenović \textit{et al.}~\cite{radenovic2018fine} propose to use 3D reconstructed models to guide the selection of the training data, and propose a novel GeM pooling layer.
Based on NetVLAD, Hausler \textit{et al.}~\cite{hausler2021patch} propose Patch-NetVLAD which combines the strengths of local and global features.
Arcanjo \textit{et al}~\cite{arcanjo2022efficient} design a lightweight network and a voting mechanism to achieve respectable results with low memory usage.


\subsection{LiDAR-based Place Recognition}
\label{sec:LPR}
The active-measurement nature of LiDAR sensors makes LiDAR-based place recognition (LPR) invariant to conditional changes such as illumination, weather and seasons. M2DP proposed by He \textit{et al.}~\cite{he2016m2dp} first projects the point cloud to multiple planes and then calculates signatures by the distributions of the projected points on different planes. Scan Context proposed by Kim \textit{et al.}~\cite{kim2018scan} divides the point cloud into sectors in polar coordinates and encodes the maximum height of laser points in each sector to generate the global descriptor matrix.
The successful application of deep learning in processing 3D point clouds has led to more and more learning-based LPR methods. PointNetVLAD by Uy \textit{et al.}~\cite{uy2018pointnetvlad} builds upon the basic functions of PointNet~\cite{qi2017pointnet} and NetVLAD~\cite{arandjelovic2016netvlad} for large-scale place recognition. After that, LPDNet proposed by Liu \textit{et al.}~\cite{liu2019lpd} combines 
 learning-based and hand-crafted local features to enhance the vanilla PointNet architecture. PCAN by Zhang~\cite{zhang2019pcan} utilizes the attention mechanism to enforce the network to pay more attention to the task-relevant regions of the environment. More recently, different input forms of point clouds have also been studied and applied to LPR tasks. Komorowski \textit{et al.}~\cite{komorowski2021minkloc3d} utilize sparse voxelized representation of point clouds as the input to the network. OverlapTransformer proposed by Ma \textit{et al.}~\cite{ma2022overlaptransformer} projects point clouds into range images to tackle the sparsity and disorder of point clouds and introduces the transformer network~\cite{vaswani2017attention} to achieve high efficiency and yaw-rotation invariance.

\subsection{Fusion-based Place Recognition}
\label{sec:FPR}
Recently, multi-sensor fusion has aroused great interest in the place recognition community. The combination of different sensors can overcome the disadvantages of the unimodal methods. 
To collaborate point clouds and images, PIC-Net~\cite{lu2020pic} introduces global channel attention to fuse features from different modalities. The Spatial Attention VLAD layer is proposed to select the discriminative points and pixels. Xie \textit{et al.}~\cite{xie2020large} use PointNet with a trimmed VLAD layer and ResNet-50~\cite{he2016deep} to extract LiDAR and visual descriptors separately, which are further fused by FC layers. CORAL~\cite{pan2021coral} by Pan \textit{et al.} constructs an elevation map from a point cloud and projects the image feature to the elevation map to semantically ``colorize" the structural features and generate a bi-modal descriptor in the bird's eye view (BEV) frame. MinkLoc++ proposed by Komorowski \textit{et al.}~\cite{komorowski2021minkloc++} uses 3D sparse convolutions to extract spatial features from point clouds, and uses pre-trained ResNet blocks~\cite{he2016deep} for image features. The two descriptors are then concatenated along the channel dimension to form a multimodal descriptor. MMDF by Yu \textit{et al.}~\cite{yu2022mmdf} uses image semantic features to enhance point cloud features. Pointwise image features are first extracted using the method described in~\cite{vora2020pointpainting}, then point cloud features and pointwise image features are aggregated by a convolution layer. To better exploit the discriminativeness of the descriptors from different modalities, Lai \textit{et al.}~\cite{lai2022adafusion} (AdaFusion) design an attention branch in their network to weight modalities adaptively.
\begin{figure*}[ht]
  \centering
  \includegraphics[width=1\linewidth]{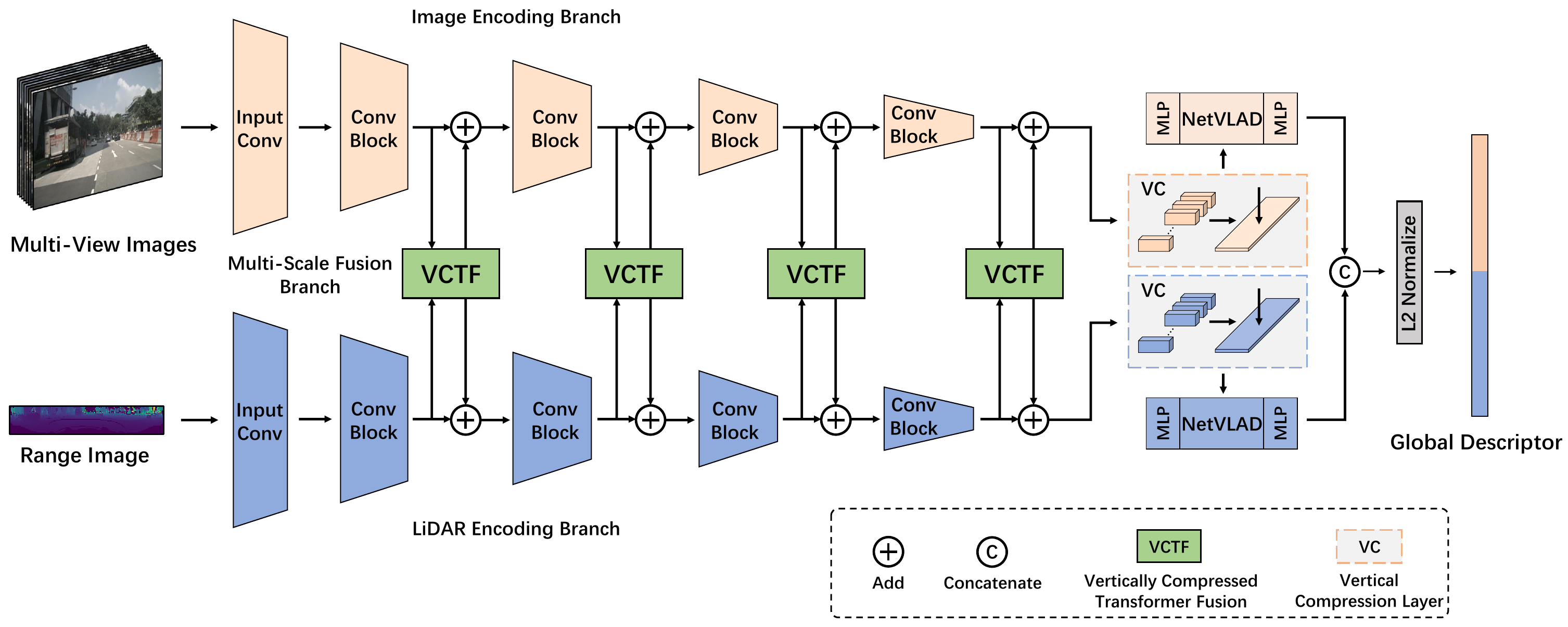}
  \caption{The overall architecture of LCPR. Multi-view RGB images and one range image are fed into the sibling image encoding (IE) branch and the LiDAR encoding (LE) branch to obtain intermediate features at different resolutions. A set of Vertically Compressed Transformer Fusion (VCTF) modules are employed to fuse these intermediate features at multiple scales. The outputs of the IE and LE branches then pass through the Vertical Compression (VC) layers to obtain the denser panoramic features, which are aggregated using NetVLAD-MLPs combos to generate sub-descriptors. Finally, the global multimodal descriptor is generated by the concatenation of the sub-descriptors, which is further used as a query or reference in the database.}
  \label{fig:overall}
\end{figure*}

Based on the aforementioned review, none of these methods fully exploit the information from different sensors to find the explicit spatial correspondence and inner feature correlations between 2D and 3D data simultaneously. In addition, the limited FOV of a single monocular camera can lead to false recognition as the viewpoint changes. Therefore, in this paper, we propose a multi-scale attention-based place recognition method that takes multi-view RGB images and range images from LiDAR scans as input to cover the panoramic view of the environment. We utilize self-attention mechanism to extract the global panoramic context between and within different modalities to generate discriminative and yaw-rotation invariant descriptors.

\section{Our Approach}
\label{sec:method}

\begin{figure*}[h]
  \centering
  \includegraphics[width=1\linewidth]{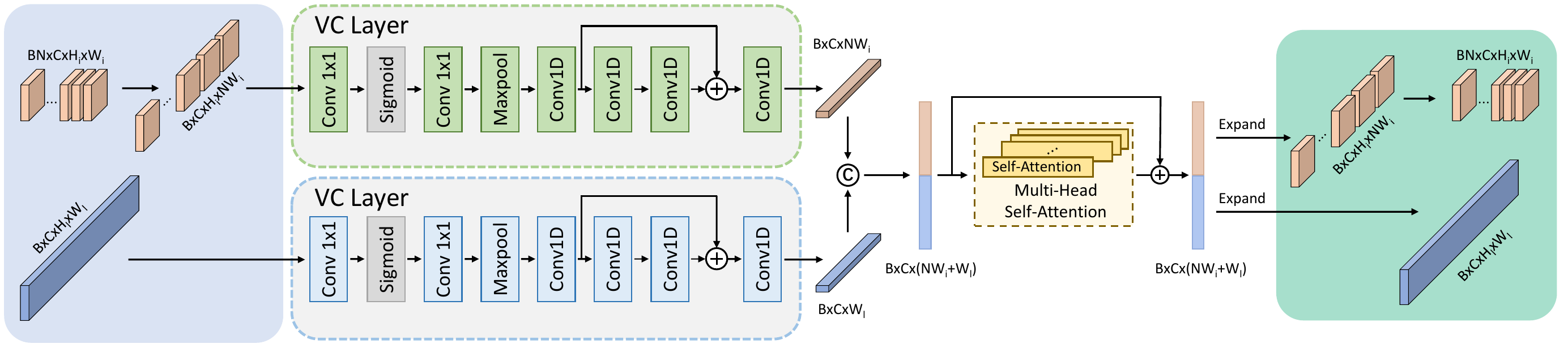}
  \caption{The architecture of Vertically Compressed Transformer Fusion (VCTF) module. The intermediate feature volumes from the two sibling feature streams are first compressed in the vertical direction. The compressed sentence-like features are concatenated horizontally, and fed into the MHSA module for multimodal fusion. The fused feature is then split and expanded to be sent back to the original feature streams.}
  \label{fig:vctf}
\end{figure*}
\subsection{Overall Architecture}
\label{sec:overall_arch}
The overall architecture of our proposed LCPR is depicted in~\figref{fig:overall}. To address the information imbalance between different modalities of limited FOV cameras and the LiDAR mounted on an autonomous vehicle, our method takes multi-view RGB images and the range image projected from point clouds as input and directly outputs the environmental global descriptor. Inspired by~\cite{prakash2021multi}, our LCPR mainly consists of an image encoding (IE) branch, a LiDAR encoding (LE) branch, a multi-scale fusion branch, and NetVLAD-MLPs combos~\cite{arandjelovic2016netvlad}.
Given the effectiveness of residual structures in processing image features~\cite{he2016deep, lu2020pic, xie2020large}, we propose to use residual blocks~\cite{he2016deep} as the feature extractors in both the IE and LE branches.

To tackle the sparsity and disorder of the point clouds, our method first produces range images using spherical projection on LiDAR point clouds. In particular, the correspondence between a LiDAR point $\bm{p}=(x,y,z)^T \in \mathbb{R}^3$ from the point cloud $\mathcal{P}$ and the pixel coordinates $(u,v)^T$ of its corresponding range image $\mathcal{R}$ can be formulated as
\begin{align}
  \left( \begin{array}{c} u \vspace{0.5em}\\ v \end{array}\right) & = \left(\begin{array}{cc} \frac{1}{2}\left[1-\arctan(y, x) \pi^{-1}\right] w   \vspace{0.5em} \\
      \left[1 - \left(\arcsin(z r^{-1}) + \mathrm{f}_{\mathrm{up}}\right) \mathrm{f}^{-1}\right] h\end{array} \right), \label{eq:projection}
\end{align}
where $r=||\bm{p}||_2$ is the range measurement of the LiDAR point, $(h,w)$ represents the height and width of the range image, and $\mathrm{f}=\mathrm{f}_{\mathrm{up}}+\mathrm{f}_{\mathrm{down}}$ is the vertical FOV of the sensor.

The multi-view RGB images and the LiDAR range images are fed into the two encoding branches, the IE branch and the LE branch, respectively. Considering that the low-level features of the network can capture rich texture information, while the high-level features can represent semantic information, we propose and leverage a set of Vertically Compressed Transformer Fusion (VCTF) modules to fuse the intermediate features from both the IE branch and the LE branch at multiple resolutions. The VCTF module employs the self-attention mechanism of transformers~\cite{vaswani2017attention} to fully incorporate the global panoramic context between and within different modalities. Subsequently, the output of the VCTF module is sent back to the encoding branches, added to the original intermediate feature, and then fed to the next residual block. The outputs of the two encoding branches are each processed by the proposed Vertical Compression (VC) layers respectively, which compress the features in the vertical direction. The compressed features are further aggregated using NetVLAD layers with MLPs to yield two sub-descriptors.
Finally, the sub-descriptors from two encoding branches are then concatenated to produce the multimodal global descriptor, which is then used for database creation and fast place retrieval.

\subsection{Vertically Compressed Multimodal Fusion}
\label{sec:VCMF}

It is argued that the height dimension contains less information than the width dimension in urban autonomous driving~\cite{zhou2023matrixvt}. Therefore, we adopt a vertical compression operation along the height dimension of the intermediate features in the pipeline to reduce the computational cost as well as to achieve yaw-rotation invariance. Here we extend the ideas of the previous work~\cite{ma2022overlaptransformer} to the multimodal fusion framework.

The architecture of our proposed VCTF module is depicted in~\figref{fig:vctf}. We denote the input feature volumes from the IE branch and the LE branch as $F_i^{in}\in \mathbb{R}^{BN \times C\times H_i\times W_i}$ and $F_l^{in} \in \mathbb{R}^{B \times C \times H_l \times W_l}$ respectively, where $N$ is the number of cameras, and $B$ is the batch size. $F_i^{in}$ is first concatenated panoramically for feature alignment, leading to $F_i^p \in \mathbb{R}^{B\times C \times H_i \times NW_i}$. Then two VC layers are designed to compress $F_i^p$ and $F_l^{in}$ in the vertical direction, respectively.
The VC layer consists of two $1\times 1$ convolutions with Sigmoid nonlinearity, a MaxPooling layer for feature compression in the vertical direction, and 1-D convolutions with Batch Normalization and ReLU nonlinearity for feature refinement.
The skip connection between the first and the third 1-D convolutions is employed to speed up the convergence.

The output feature volumes of the VC layers with sizes of $B\times C \times NW_i$ and $B \times C \times W_l$ are concatenated along the width dimension to generate a sentence-like feature $F_s\in \mathbb{R}^{B \times C \times (NW_i+W_l)}$. And then we feed $F_s$ into a multi-head self-attention (MHSA) module~\cite{vaswani2017attention} to extract the correlations between modalities and the global context within each modality simultaneously. The self-attention mechanism can be formulated as
\begin{align}
    \mathcal{A} = \mathrm{Attention} (Q,K,V)=\mathrm{softmax}\left ( \frac{QK^T}{\sqrt{d_k} }  \right )V,
    \label{eq:SA}
\end{align}
where $Q$, $K$, $V$ represent the queries, keys, and values respectively, $d_k$ is the dimension of keys.

Finally, the output feature volumes of the MHSA module are expanded along the height dimension, so the inputs and outputs of the VC layers have the same size for the fusion with the original intermediate features.

\subsection{Yaw-Rotation Invariance}
\label{sec:yaw_rotation_invariance}
Our proposed LCPR is designed to generate yaw-rotation invariant global descriptors, ensuring the robustness of our method in viewpoint changes in real-world autonomous driving applications. In this section, we provide a comprehensive mathematical derivation of the yaw-rotation invariance achieved by our network.
It has been reported in~\cite{uy2018pointnetvlad, ma2022overlaptransformer, ma2022seqot, ma2023cvtnet} that the NetVLAD module is permutation-invariant. Hence, here we concentrate on exploring the yaw-rotation equivariance of the modules before NetVLAD in our LCPR.

\textit{1) Residual Unit}: The residual unit in our IE branch and LE branch can be defined as
\begin{align}
    \mathbf{y}=\sigma \left ( W_s\mathbf{X}+ W_2\sigma \left ( W_1 \mathbf{X}  \right )  \right ).
\end{align}
Here $\mathbf{X}$ and $\mathbf{y}$ are the input and output feature volumes, $\sigma \left ( \cdot\right )$ denotes the ReLU function, $W_1$ and $W_2$ represent two convolution layers, and $W_s$ represents a linear projection for matching dimensions.

Then we right-multiply the input $\mathbf{X}$ with a matrix $C$ to represent a column shift corresponding to a yaw-angle rotation of the vehicle. Since the convolution operation and ReLU function are both translation equivariant, we have
\begin{equation}
    \begin{aligned}
    \mathbf{y}_s & = \sigma \left ( W_s(\mathbf{X}C)+W_2\sigma \left ( W_1 (\mathbf{X}C)  \right )  \right )\\
    &=\sigma \left ( W_s\mathbf{X}C+W_2\sigma \left ( W_1 \mathbf{X}  \right )C  \right )\\
    &=\sigma \left ( W_s\mathbf{X}+W_2\sigma \left ( W_1 \mathbf{X}  \right )  \right )C\\
    &=\mathbf{y}C.
    \end{aligned}
\end{equation}

As can be seen, the residual unit is translation equivariant, resulting in the two yaw-rotation equivariant encoding branches.

\textit{2) VCTF module}:
The VCTF module mainly consists of two parts: the VC layer and the MHSA module. As described in ~\secref{sec:VCMF}, the VC layer has a fully convolutional structure, thus possessing yaw-rotation equivariance.

As for the MHSA module, we denote the vertically compressed feature volumes extracted by two VC layers as $\mathbf{X}_i \in \mathbb{R}^{D\times NW_i}$ and $\mathbf{X}_l \in \mathbb{R}^{D \times W_l}$. To avoid ambiguity, here we use $D$ to represent the feature dimension. Then the input of the MHSA module can be denoted as $\mathbf{X} = \left [  \mathbf{X}_i \; \mathbf{X}_l\right ]^T $, which represents the concatenation of the two features along the width dimension. With column shift matrices $C_i$ and $C_l$, the input becomes
\begin{equation}
    \begin{aligned}
    \mathbf{X}_s&=\begin{bmatrix}\mathbf{X}_iC_i &\mathbf{X}_lC_l \end{bmatrix}^T\\ & = \begin{bmatrix}C_i^T
    &\mathbf{0} \\\mathbf{0}
    &C_l^T
    \end{bmatrix}\begin{bmatrix}\mathbf{X}_i
    &\mathbf{X}_l
    \end{bmatrix}^T\\
    &=C\mathbf{X}.
    \end{aligned}
\end{equation}
Here $C \in \mathbb{R}^{2D\times 2D}$ represents the transformation matrix of the input $\mathbf{X}$ resulting from the column shifts of the features $\mathbf{X}_i$ and $\mathbf{X}_l$.
Then~\eqref{eq:SA} becomes
\begin{equation}
    \begin{aligned}
    \mathcal{A}_s &=\mathrm{Attention}(Q_s,K_s,V_s)  = \mathrm{softmax}\left (  \frac{Q_sK_s^T}{\sqrt{d_k} }  \right ) V_s\\
   &=\mathrm{softmax}\left (\frac{\left (C\mathbf{X}W_Q \right )\left ( {W_K}^T\mathbf{X}^TC^T \right )}{\sqrt{d_k}}   \right ) C\mathbf{X}W_V \\
   &=C\,\mathrm{softmax}\left ( \frac{\left (\mathbf{X}W_Q \right )\left ( {W_K}^T\mathbf{X}^T \right )}{\sqrt{d_k}} \right ) C^TC\mathbf{X}W_V\\
   &=C\,\mathrm{softmax}\left ( \frac{QK^T}{\sqrt{d_k} } \right )V \\
   &=\begin{bmatrix}\mathbf{Y}_iC_i &\mathbf{Y}_lC_l\end{bmatrix}^T,
    \end{aligned}
\end{equation}
where $\mathbf{Y}_i$ and $\mathbf{Y}_l$ are the output feature volumes of MHSA corresponding to the inputs $\mathbf{X}_i$ and $\mathbf{X}_l$, respectively.

As can be seen, the column shifts $C_i$ and $C_l$ are preserved from the input to the output, showing that the MHSA module is yaw-rotation equivariant. Consequently, the VCTF module is proven to be yaw-rotation equivariant, leading to the holistic LCPR yaw-rotation equivariant.

\begin{table}[t]
  \centering
  \begin{center}
  	\renewcommand\arraystretch{1.4}
    \caption{Statistics of our data organization}
        \footnotesize{
        \begin{tabular}{c|ccc}
          \toprule
          Location & Boston-Seaport & SG-OneNorth & SG-HollandVillage \\ \hline
          N$_\text{sample}$ & 22103 & 8104 & 3427 \\ \hline
          N$_\text{database}$ & 9686 & 4796 & 2433 \\ \hline
          N$_\text{query\_train}$ & 8407 & -- & -- \\ \hline
          N$_\text{query\_val}$  & 1002 & -- & -- \\ \hline
          N$_\text{query\_test}$  & 3008 & 3308 & 994 \\ \hline
          Time & Day & Day & Day+Night \\
          \bottomrule
          \multicolumn{4}{p{0.9\linewidth}}{N$_\text{sample}$, N$_\text{database}$, N$_\text{query\_train}$, N$_\text{query\_val}$, and N$_\text{query\_test}$ are the numbers of samples, database, training queries, validation queries, and testing queries respectively.}\\
        \end{tabular}
        }
    \label{tab:dataset}
    \end{center}
\end{table}

\begin{table*}[ht]
  \centering
  \begin{center}
  	\renewcommand\arraystretch{1.25}
    \caption{Evaluation of place recognition performance on the BS and SON data split}
    \footnotesize{
        \begin{tabular}{cccccccccc}
          \toprule
          \multirow{2}{*}{Methods} & \multirow{2}{*}{Modality$^1$} &\multicolumn{4}{c}{BS split}&\multicolumn{4}{c}{SON split} \\ \cline{3-10}
         ~&~ & Recall@1 &Recall@5 & Recall@10 & max $F_1$ & Recall@1 &Recall@5 & Recall@10 & max $F_1$ \\ \hline
          NetVLAD~\cite{arandjelovic2016netvlad} & V & 90.26 & 95.58 & 96.51 & 0.9510 & 96.25 & 98.07 & 98.58 & 0.9808\\
          PointNetVLAD~\cite{uy2018pointnetvlad} & L & 74.30 & 84.54 & 87.53 & 0.8885 & 98.49 & 99.43  &99.58 & 0.9931\\
          OverlapTransformer~\cite{ma2022overlaptransformer} & L & 82.05 & 87.67 & 88.96 & 0.9050  & 98.73 & 99.70 & 99.76 & 0.9937  \\
          AutoPlace~\cite{cait2022autoplace} & R & 90.46 & 97.27 & 98.40 & 0.9499 & 92.99 & 97.85 & 98.79 & 0.9636 \\
          AdaFusion~\cite{lai2022adafusion} & V+L & 86.74 & 95.58 & 97.17 & 0.93187 & 98.70 & 99.58 & \textbf{99.94} & 0.9942 \\
          PIC-Net~\cite{lu2020pic} & V+L & 88.36 & 95.21 & 97.17 & 0.9445 & 98.13 & 98.79 & 99.06 & 0.9905 \\
          MinkLoc++~\cite{komorowski2021minkloc++} & V+L & 77.76 & 90.43 & 93.68 & 0.8753 & 91.32 & 97.28 & 98.61 & 0.9556 \\
          LCPR (ours) & V+L & \textbf{97.34} & \textbf{99.24} & \textbf{99.50} & \textbf{0.9873} & \textbf{99.40} & \textbf{99.79} & 99.88 & \textbf{0.9969}  \\
          \bottomrule
        \multicolumn{10}{p{0.9\linewidth}}{$^1$ V: Visual, L: LiDAR, V+L: Visual+LiDAR.}\\
        \end{tabular}
        }
    \label{tab:bs_son}
    \end{center}
\end{table*}


\subsection{Network Training}
\label{sec:network_training}
We follow~\cite{arandjelovic2016netvlad} and~\cite{cait2022autoplace} to use the triplet margin loss to train the network. Specifically, for each training step, we construct a min-batch $\left( q, p^q, \{n_i^q \}\right)$ containing a query, a positive sample, and several negative samples. Following~\cite{cait2022autoplace}, we define a positive sample as one that is located no more than 9\,meters apart from the capture location of the query. Conversely, we define a sample as negative if it is at least 18\,meters apart from the capture location of the query.

We denote $f(\cdot)$ as the mapping from the input to its global descriptor. Our goal is to minimize the distance between $f(q)$ and $f(p^q)$ while maximizing the distance between $f(q)$ and $f(n_i^q)$. It has been reported that the network is prone to overfitting in the image domain during training~\cite{komorowski2021minkloc++}. To mitigate this issue, for each training query, we select a positive sample with the smallest distance between $f(p_i^q)$ and $f(q)$ from the potential positive group to form the mini-batch. The negative samples are randomly selected from the potential negative group. In addition, to increase the training efficiency, we only retain negative samples that satisfy
\begin{align}
    d\left ( f(q), f(n_i^q)\right )<margin.
\end{align}

Then, the loss function is given by
\begin{align}
    \mathcal{L}_T=\frac{1}{N_\text{neg}} \sum_{i=1}^{N_\text{neg}} \left [ d\left ( f(q),f(p^q) \right )-d\left ( f(q), f(n_i^q) \right )+m \right ] _+,
\end{align}
where $\left [ \cdots \right ]_+$ denotes the hinge loss, $d(\cdot)$ is the Euclidean distance, $m$ is a constant giving the margin, and $N_\text{neg}$ represents the number of the selected negative samples.

\section{Experiments}
\label{sec:experiments}


\subsection{Experimental Setup and Implementation Details}
\label{sec:expe_setup}
Since our method requires both multi-view RGB images with $360^{\circ}$ horizontal FOV and LiDAR point clouds as input, we evaluate our method on the nuScenes dataset~\cite{caesar2020nuscenes}, which comprises 1000 driving scenes, each with a duration of 20 seconds, and carries the full autonomous vehicle sensor suite including six cameras, one LiDAR, and five radars.
The dataset is collected from four different locations in Boston and Singapore: Boston Seaport (BS), SG-OneNorth (SON), SG-Queenstown (SQ), and SG-HollandVillage (SHV). Among them, the BS split is the largest, containing 550 scenes, and the others contain 215, 135, and 100 scenes respectively. Due to the limited number of samples and the monotonous environment of the SQ split, we only conduct experiments on the BS, SON, and SHV splits.

We organize the data following the previous work~\cite{cait2022autoplace}. However, we introduce a modification by creating the \textit{database set} from the entire data split to ensure that the database is up-to-date. The rest of the data is used as the query set and is divided into the \textit{training query set}, the \textit{validation query set}, and the \textit{test query set}. For SON and SHV split, we only divide the data into the \textit{database set} and the \textit{test query set}. Finally, places in the query set that have no ground truth in the \textit{database set} are removed. More details of our data organization can be found in~\tabref{tab:dataset}

For our proposed LCPR, we use LiDAR range images of size $1 \times 32 \times 1056$, and each input RGB image is resized from $3 \times 1600 \times 900$ to $3 \times 704 \times 256$ resolution for data alignment between the two modalities. We use the residual blocks from ResNet-18~\cite{he2016deep} pre-trained on ImageNet~\cite{deng2009imagenet} in our IE and LE branches. For the MHSA module, we set the number of heads $n_{head}=4$ and the embedding dimension $d_k=d_v=d_{model}$. For the NetVLAD modules, we set the number of clusters $d_{cluster}=32$, the output dimension $d_{sub}=128$, generating two sub-descriptors with size of $1 \times 128$. The final multimodal descriptor then has a size of $1 \times 256$. For the network training, the ADAM optimizer is used to update the weights of the network. The initial learning rate is set to $1 \times 10^{-5}$ and decays by a factor of 0.8 every 10 epochs. For each mini-batch, we set the number of negative samples $N_{neg}=6$ and $m=0.5$ for the triplet loss.

In the following experiments, we compare the performance of our LCPR with the state-of-the-art baselines including the vision-based method NetVLAD~\cite{arandjelovic2016netvlad}, the radar-based method AutoPlace~\cite{cait2022autoplace}, the LiDAR-based methods PointNetVLAD~\cite{uy2018pointnetvlad} and OverlapTransformer~\cite{ma2022overlaptransformer}, and multimodal methods including MinkLoc++~\cite{komorowski2021minkloc++}, PIC-Net~\cite{lu2020pic}, and AdaFusion~\cite{lai2022adafusion}.
We use the released open-source codes of the baseline methods for evaluation except for PIC-Net~\cite{lu2020pic}, which is implemented ourselves according to their original paper. All experiments are conducted on a system with an Intel i5-12400F CPU and an Nvidia RTX 3080 GPU.

\subsection{Evaluation for Place Recognition}
\label{sec:eval_res}
In order to evaluate the place recognition performance of our proposed method, we first train and test LCPR as well as other state-of-the-art methods on the largest BS split. Furthermore, we also conduct a generalization study with the SON split on all the methods using the weights trained on the BS split. To ensure a fair comparison, we only input single-frame data without RCS histogram Re-Ranking for AutoPlace. In our experiments, AutoPlace and NetVLAD output descriptors of size $1\times9216$, while the other baselines produce descriptors of size $1\times 256$.

Following the previous works~\cite{cait2022autoplace, ma2022seqot}, we report Recall@N and max $F_1$ scores as evaluation metrics, which are shown in~\tabref{tab:bs_son}.
First, we can see that the recalls of all the methods on the SON split are generally higher than those on the BS split. This can be attributed to the fact that the driving environments in the BS split are much more complex with richer scenes and more observation samples.
Our LCPR outperforms all other unimodal and multimodal baselines on both the splits, indicating that our method not only achieves better performance on the previously seen places after training, but also generalizes well to the unseen scenarios without fine-tuning. Although AutoPlace can achieve relatively good performance on the BS split, its zero-transfer results on the SON split are not outstanding compared to other methods, indicating its limited generalization ability. 
Furthermore, compared with AutoPlace and NetVLAD, the descriptors generated by our LCPR are more compact but more discriminative, which means our approach imposes a reduced computational burden on both query retrieval and database storage.

\subsection{Ablation Study}
\label{sec:ablation}
In this section, we investigate the effectiveness of our devised multimodal fusion mechanism. 
We compare LCPR with its two unimodal baselines, LCPR-L and LCPR-I, which only contain the image encoding branch and LiDAR encoding branch respectively.
For a fair comparison, we increase the output dimension of the NetVLAD layer to 256 in each unimodal branch.
The evaluation results are shown in the first row of~\tabref{tab:occlusion}.
Our proposed multi-scale modality fusion mechanism can effectively exploit the environmental information and further improve the place recognition performance. It can also be observed that the LiDAR range input produces better results than the visual RGB input in our network, suggesting that LiDAR features have a greater contribution to the distinctiveness of our multimodal descriptor.
This can be attributed to the RGB input's deficiency in providing depth information and the impact of raindrops on the camera lens.

\subsection{Robustness Study}
\label{sec:robustness}
We further investigate the robustness of our LCPR in multiple cases with occlusion and changing lighting conditions.

\textit{Occlusion}: In real-world environments, sensors can be occluded by raindrops and water mist that are attached to them. However, there is a lack of public datasets that contain both extreme weather conditions and visual inputs with a horizontal FOV of $360^{\circ}$. Therefore, we perform manual degradation on the nuScenes dataset to demonstrate the effectiveness of multimodal fusion in dealing with such situations. Specifically, we set all pixels of each camera image to zero in turn to simulate the camera being completely blocked.
For LiDAR, we also occlude one-quarter of the range images along the height dimension.
As shown in~\tabref{tab:occlusion}, when a certain sensor is disturbed, LCPR can still achieve about 90\% of Recall@1, while the performance of the two unimodal baselines declines. This demonstrates that LCPR using multimodal inputs has strong robustness against occlusion cases.

\begin{figure}[t]
  \centering
  \includegraphics[width=1\linewidth]{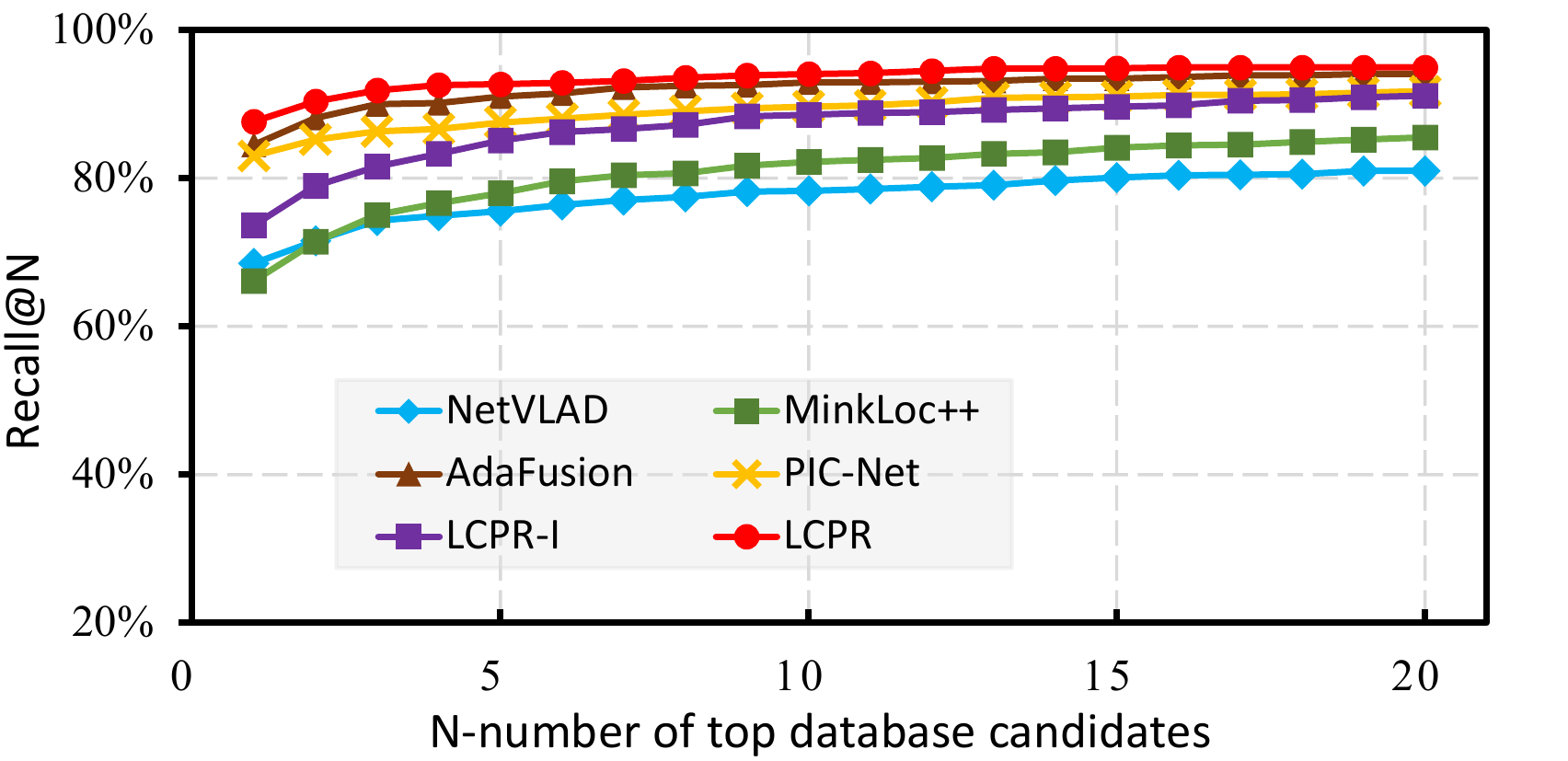}
  \caption{Place recognition results on the SHV split.}
  \label{fig:shv}
\end{figure}

\begin{figure}[t]
  \centering
  \includegraphics[width=1\linewidth]{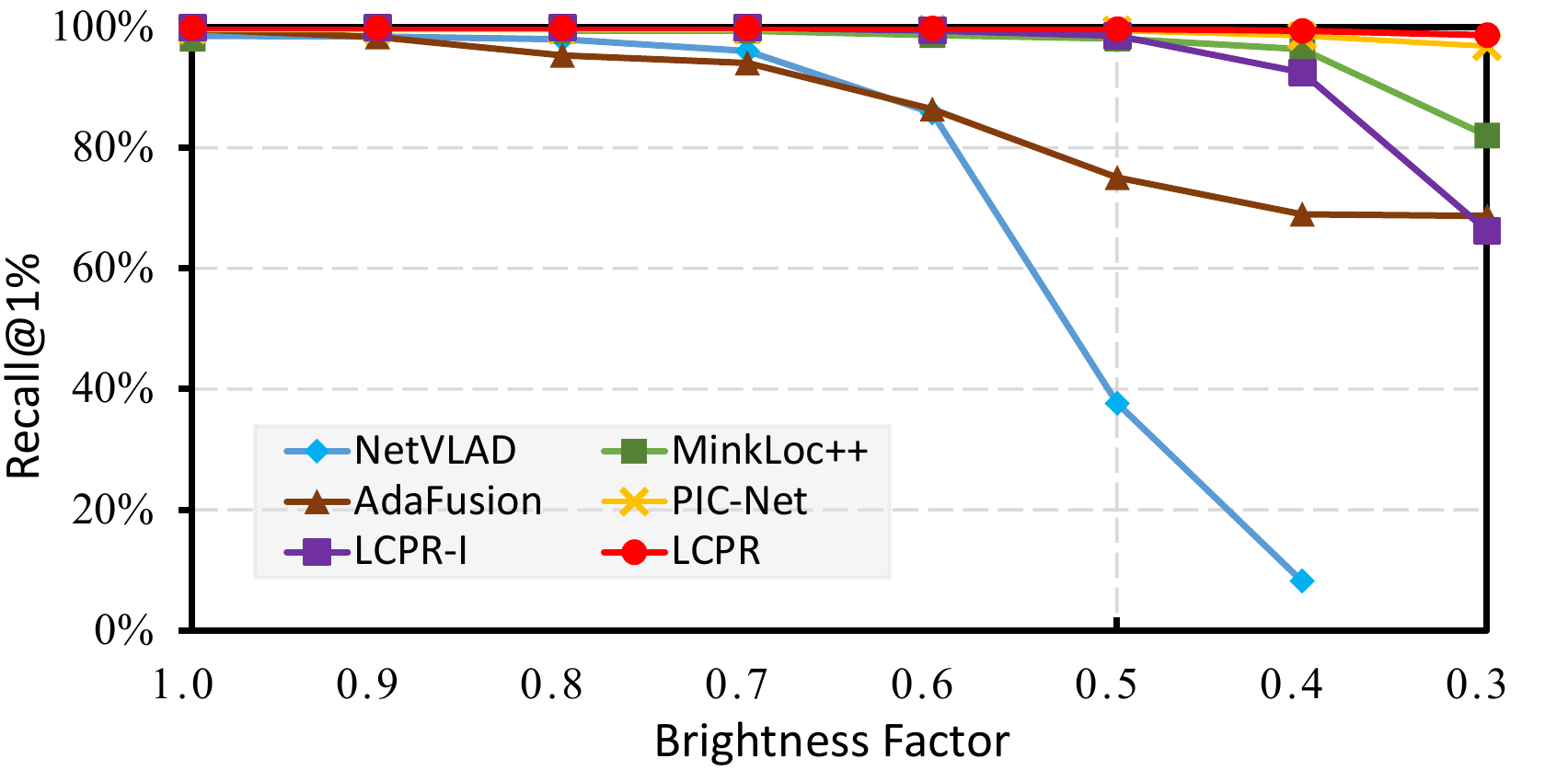}
  \caption{Place recognition results on the BS split when manually adjusting image brightness.}
  \label{fig:BS_lighting}
\end{figure}

\textit{Lighting}: Variations in lighting conditions significantly affect the visual sensors, which is one of the starting points for introducing multimodal fusion in place recognition. Therefore, to show the robustness of our method in tackling changing lighting conditions, we also evaluate LCPR against baselines that also incorporate visual information of the environment on the SHV and BS split. The SHV split naturally contains many day-to-night repetitive driving trajectories. For the BS split, we manually reduce the image brightness in the query sample with a series of brightness factors to simulate changes in lighting conditions. As shown in~\figref{fig:shv} and~\figref{fig:BS_lighting}, multimodal methods generally have better performance than the unimodal method, i.e., NetVLAD with only visual input. The results demonstrate that incorporating the LiDAR point cloud can mitigate the impact of changes in lighting conditions on place recognition tasks. Note that our LCPR still achieves the best performance, indicating its strong robustness against natural lighting condition changes and manual brightness adjustment.

\begin{table}
    \centering
    \tabcolsep=0.3cm
    \renewcommand\arraystretch{1.15}
    \caption{Occlusion test results (Recall@1) on the BS split}
    \begin{tabular}{c|ccc}
        \toprule
         Occluded Sensor & LCPR & LCPR-I & LCPR-L \\ \hline
         -- & \textbf{97.34} & 93.88 & 96.21 \\
         CAM\_FRONT & \textbf{89.06} & 81.35 & -- \\
         CAM\_FRONT\_RIGHT & \textbf{88.26} & 86.84 & --  \\
         CAM\_BACK\_RIGHT & \textbf{87.53} & 83.71 & -- \\
         CAM\_BACK & \textbf{91.02} & 82.85 & --  \\
         CAM\_BACK\_LEFT & \textbf{90.26} & 84.21 & --  \\
         CAM\_FRONT\_LEFT & \textbf{88.66} & 87.00 & -- \\
         LIDAR\_TOP & \textbf{88.66} & -- & 64.69 \\      
         \bottomrule
    \end{tabular} 
    \label{tab:occlusion}
\end{table}

\begin{figure}
  \centering
  \includegraphics[width=1\linewidth]{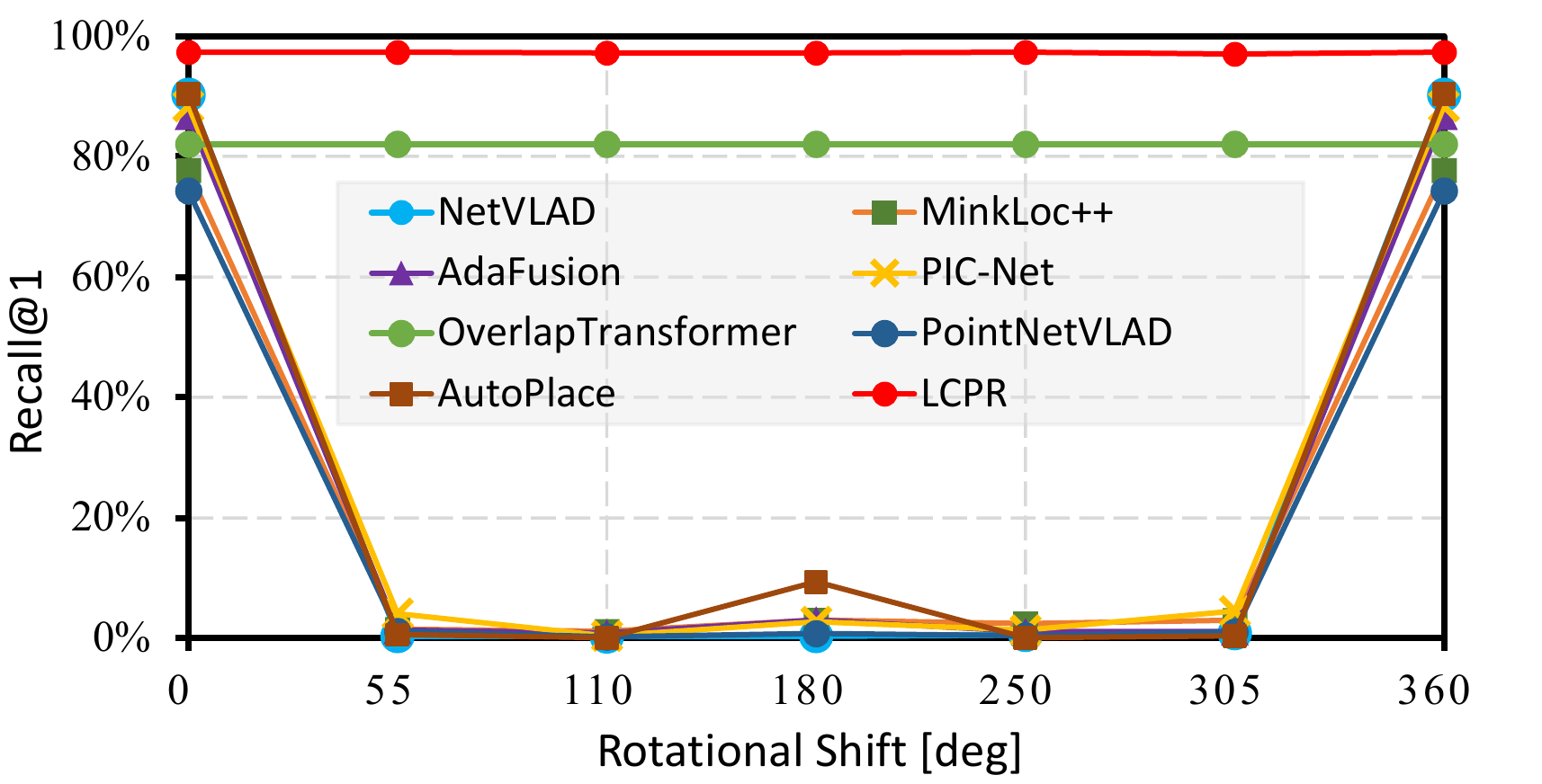}
  \caption{Study on yaw-rotation invariance.}
  \label{fig:yaw}
\end{figure}

\subsection{Yaw-Rotation Invariance Study}
\label{sec:yaw_invariance}
In this experiment, we investigate the yaw-rotation invariance of our proposed descriptor on the BS split. Similar experiment pipelines can be found in the previous works~\cite{ma2022overlaptransformer, ma2022seqot, ma2023cvtnet}, which only focus on the LiDAR-based place recognition. Given that the nuScenes dataset provides multi-view RGB images covering a $360^{\circ}$ horizontal perspective, we can extend the pipeline to the multimodal field.
For visual input, we first concatenate images of the same frame in the width direction and subsequently apply a circular shift to approximate the yaw-rotation of the vehicle.
For point cloud input, we directly rotate the point cloud around the yaw-axis.
Considering the shooting angle of the images in the nuScenes dataset, we rotate each query sample by $\left [ 55^{\circ},110^{\circ},180^{\circ},250^{\circ},305^{\circ},360^{\circ} \right ]$ while searching for the places with respect to the same original database.
We use Recall@1 as the evaluation metric in this experiment.
As illustrated in~\figref{fig:yaw}, both LCPR and OverlapTransformer are not affected by rotation along the yaw-axis, while other baseline methods show a significant drop in performance when a yaw-rotation is applied to the query. Although OverlapTransformer is also yaw-rotation invariant, it still has a lower Recall@1 compared to our LCPR.

\subsection{Runtime and Memory Consumption}
\label{sec:runtime}
In this section, we conduct the evaluation of LCPR's real-time performance and memory usage. 
The system configuration of this experiment has been clarified in~\secref{sec:expe_setup}. 
For each query, we accumulate the time consumed in generating the descriptor and retrieve the top-20 candidates from the database to calculate the average runtime. Note that the first 100 queries are used as a warm-up, which are not included in the average runtime. Here we use FAISS library~\cite{johnson2019billion} to improve the search efficiency. LCPR takes an average of 16.38\,ms to generate a descriptor (12.17\,ms) and select the top-20 candidates (4.21\,ms) from the database of 9686 samples of the BS split, showing that our method can operate online in real-vehicle applications.
Furthermore, our LCPR has a parameter count of 28.97\,M, which also indicates that our method has a low memory consumption.

\section{Conclusion}
\label{sec:conclusion}
In this paper, we present LCPR, a novel and generic multimodal fusion network for place recognition. 
Our method takes multi-view RGB images from cameras and range images from LiDAR scans as input, and generates global descriptors that are robust to lighting conditions and viewpoint changes.
Benefiting from the self-attention mechanism of the transformer structure, our network can fully exploit the correlations between modalities and the global context within each modality simultaneously, and integrate representations of different modalities. Experimental results show that our descriptor outperforms state-of-the-art unimodal and multimodal methods on the publicly available dataset and can operate online in real-vehicle applications.


\ifCLASSOPTIONcaptionsoff
  \newpage
\fi

\bibliographystyle{IEEEtran}

\bibliography{IEEEabrv,ref}

\end{document}